\documentclass[12pt]{article}

\usepackage{times}  
\usepackage{helvet} 
\usepackage{courier}  
\usepackage[hyphens]{url}  
\usepackage{graphicx} 
\usepackage{graphicx}  
\usepackage[round]{natbib}
\newenvironment{sciabstract}{%
\begin{quote} \baselineskip14pt\small\hfil {\bf Abstract} \hfil\\[3pt]}
{\end{quote}\vspace{6pt}}

\renewcommand{\cite}{\citep}

\usepackage{flushend}

\usepackage{amsmath}
\DeclareMathOperator*{\argmax}{arg\,max}
\DeclareMathOperator*{\argmin}{arg\,min}
\usepackage{amsfonts}
\usepackage{booktabs}
\usepackage{algorithm}
\usepackage{algorithmicx}
\usepackage{algpseudocode}
\usepackage[export]{adjustbox}
\usepackage{subfigure}
\usepackage{longtable}
\usepackage{tabularx}

\title{Improving Fictitious Play Reinforcement Learning with Expanding Models}

\author
{Rong-Jun Qin$^{1,2}$, Jing-Cheng Pang$^1$, Yang Yu$^{1,\dag}$\\
\normalsize{$^1$National Key Laboratory for Novel Software Technology, Nanjing University, China}\\
\normalsize{$^2$Polixir}\\
\normalsize{emails: qinrj@polixir.ai, pangjc@lamda.nju.edu.cn, yuy@nju.edu.cn.}\\
\normalsize{$^\dag$To whom correspondence should be addressed}
}

\date{}

\begin{document}

\baselineskip16pt

\maketitle 

\begin{sciabstract}
Fictitious play with reinforcement learning is a general and effective framework for zero-sum games. However, using the current deep neural network models, the implementation of fictitious play faces crucial challenges. Neural network model training employs gradient descent approaches to update all connection weights, and thus is easy to forget the old opponents after training to beat the new opponents. Existing approaches often maintain a pool of historical policy models to avoid the forgetting. However, learning to beat a  pool in stochastic games, i.e., a wide distribution over policy models, is either sample-consuming or insufficient to exploit all models with limited amount of samples. In this paper, we propose a learning process with neural fictitious play to alleviate the above issues. We train a single model as our policy model, which consists of sub-models and a selector. Everytime facing a new opponent, the model is expanded by adding a new sub-model, where only the new sub-model is updated instead of the whole model. At the same time, the selector is also updated to mix up the new sub-model with the previous ones at the state-level, so that the model is maintained as a behavior strategy instead of a wide distribution over policy models. Experiments on Kuhn poker, a grid-world Treasure Hunting game, and Mini-RTS environments show that the proposed approach alleviates the forgetting problem, and consequently improves the learning efficiency and the robustness of neural fictitious play.
\end{sciabstract}

\maketitle

\section{Introduction}
Many multi-agent systems can be modelled by games with incomplete information, where players have only partial observations of the system states and take actions based on these observations. Usually, the goal of each player is to maximize their expected payoff locally. However, it is intractable to obtain a closed-form solution based on linear programming or some search techniques as the observation or action space increases. An alternative approach is learning in games where the player learns a strategy through playing the game. Fictitious play \cite{brown1951fp} was a famous learning framework in games. In each iteration of fictitious play, the player makes a best response ccording to their belief about the other players' strategies, repeatedly plays the game and averages these best responses. In specific games, the average strategy profile can be shown to converge to a Nash equilibrium.

There are many variants of standard fictitious play \cite{fudenberg1998learning}, but most of them focus on normal-form representations and cannot be readily applied in large-scale games. On the other hand, although reinforcement learning (RL) is an efficient approach to solve large-scale single-agent sequential decision-making problems, vanilla independent reinforcement learning often fails in multi-agent scenarios with new challenges mainly from two aspects: \emph{instability} and \emph{adaptability}  \cite{marl_survey08}. The simultaneously indepent learning process makes the learning dynamics of every agent instable, and the learning agent must adapt to the changing behaviors of other agents. To our knowledge, FSP \cite{heinrich15fsp} was the first to successfully apply fictitious play in large-scale imperfect-information extensive-form games with a general machine learning framework, without any explicit knowledge about the opponents or the game. In FSP, RL was used for solving best response, and the average strategy was learned by supervised learning from best response experiences. An improved version with neural networks named NFSP \cite{16nfsp} was also proposed.

Despite the success of reinforcement learning in games, there are crucial issues with current deep RL techniques that need to be solved. Avoiding forgetting the best responses to old opponents is essential when competing with a possibly dynamic opponent, e.g., an opponent that also learns. However, the most frequently used neural function approximations are easy to overfit to the current opponent and have an unknown effect on previously well-trained weights due to the back-propagation updates. Existing approaches try to finalize these trained policy models, i.e., maintain a pool of historical policy models and a distribution over these models.

In this paper, we focus on decentralized learning in partially observable environments, and propose a fictitious play reinforcement learning framework with expanding models named FPEM to address the above issues. In every iteration, the FPEM expands with a sub-model, which is also a neural network policy that is a best response to the opponent. Then a policy selector network mixes all these sub-models, updates during training and chooses a sub-model at every state. Intuitively speaking, these sub-models act as base policies and try to handle diverse malicious opponents, while the selector network handles in an upper level to choose a better candidate base policy at every state.

\noindent\textbf{Related Works.} \indent CFR \cite{07cfr} appeared to be the first tractable approach for finding the Nash equilibrium in large-scale imperfect-information extensive-form games through learning in games. CFR and its variants, such as CFR+ \cite{14cfrplus}, linear CFR \cite{linear_cfr}, performed well in practice. CFR-based approaches usually need to traverse the game tree to reason about every state and require some knowledge to abstract the game. Recently, Deep CFR \cite{deep_cfr19} was also proposed to obviate the need for abstratction and using neural networks to deal with large games and comparable to linear CFR. Deepstack \cite{2017deepstack} also incorporated reinforcement learning, and was capable of beating professional poker players at heads-up no-limit Texas hold'em poker based on CFR.

Double Oracle (DO) algorithm \cite{DO03} maintained two deterministic policy pool in two-player games and assumed each player is restricted to select from their policy pool, then DO algorithm iteratively found an optimal pure strategy for each player against its opponents and added them to the two pools. Policy space response oracle (PSRO) \cite{PSRO} generalized DO, where the meta-game's choices are policies rather than pure policies. The PSRO then learns the combination model (meta-strategy) of the oracles from the expected utility tensor computed via simulations. Based on the learning objective of meta-strategy, PSRO can be instances of independent RL, FP and DO. Functional-form games and gamespaces were proposed to construct a sequence of objectives \cite{PSRO_rn} and a rectified RSPO algorithm was also proposed. 

To balance the performance of learning agents in two-player stochastic games, soft Q-learning was applied with a uniform distribution regularization term, to design games with adaptable and balancing properties tailored to human-level performance \cite{ijcai18bsql}. Previous work \cite{aistats18acfp} built a stochastic approximation of the fictitious play process in an online, decentralized fashion and applied it to multistage games, where the best response model chooses an action to maximize a perturbed Q function. In partially observable multi-agent environments, the actor-critic framework with several policy update rules based on regret minimization can lead to previously unknown convergence guarantees \cite{nips18acfp} in such environments. An variant of NFSP is applied to an RTS game, replacing the best response solver with PPO \cite{nfsp_rts19}, and the authors launched multiple processes to reduce off-policy data. Exploitability Descend \cite{ijcai19_ED} computed a best response and then performed exploitability descend directly on the policy to increase the utility, without having to compute an explicit average policy. 

\noindent\textbf{Contributions. } We note that the previous works either used one single fixed structure policy model (usually a neural network) as an average strategy and updated on that model or maintained a popopulation of policies (policy pool), evolved the pool and updated the sampling distribution over the population, based on the fitness/performance of policies in the population. Our major contributions are that we propose an expanding policy model that consists of multiple trained sub-models, and can be used as a behavior strategy, stabilizing the performance against the old opponents and also adapting to the new opponents. Combined with the opponent pool, the FPEM alleviates the forgetting problem and continually incorporates a newly trained base policy to form a new mixed model. The selector network learns from the specified distribution over these sub-models and the newly added sub-model learns to adapt to the opponent. Since the policy selector is a behavior policy, it is able to adjust base policies during one game and achieve more stable performance than sampling a policy for the whole game from the policy pool, although both approaches can achieve the same long-term utility in expectation. Besides, the FPEM model with a maximum number of base policies is more flexible and excels when compared with a single policy network with more parameters.

\section{Preliminaries}
\subsection{Reinforcement learning}
In reinforcement learning (RL) \cite{sutton2018rl}, an agent interacts with the environment, i.e., takes an action and receives a state or observation and reward signal from the environment at each timestep. The goal of the RL agent is to learn an optimal policy that maximizes the expected {cumulative rewards}, i.e., the {return} in the long run for the agent from interacting trajectories. RL can be formalized as a Markov decision process (MDP). Formally, an MDP is defined as a 5-tuple $\langle S,A,P,r,\gamma\rangle$ where the state space is $S$, with the action space $A$, and $P:S\times A\times S \rightarrow [0, 1]$ is the state transition function. Namely, when the agent takes action $a_t$ in state $s_t$, it transitions to a new state $s_{t+1}\sim P(s_{t+1}|s_t,a_t)$ and gets the reward signal $r:S\times A \rightarrow \mathbb{R}$. The last term $\gamma\in [0, 1]$ is the discount factor that trades off the current and future rewards. The agent's policy is a probability distribution over state--action space $\pi:S\times A \rightarrow [0, 1]$ satisfying $\sum_{a\in A}\pi(s,a)=1$. The state value function $V$ and state--action value function $Q$ is defined by $V^\pi(s)=\mathbb{E}^\pi[\sum_{t=0}^\infty \gamma^t r(s_{t}, a_t)|s_0=s]$ and $Q^\pi(s, a)=\mathbb{E}^\pi[\sum_{t=0}^\infty \gamma^t r(s_t, a_t)|s_0=s, a_0=a]$. With the notation of value function $V$, the goal of RL can be formulated as $\pi^\star(s,a)=\argmax_\pi V^\pi(s)=\argmax_\pi\mathbb{E}^\pi[\sum_{t=0}^\infty \gamma^t r(s_t, a_t)|s_0=s]$. In partially observable environments, the agent receives only a partial observation $o_t$ of the system state, and to circumvent the belief state in POMDP, the agent's policy maps a state-action history sequenece $h_t = (o_1, a_1, \cdots, o_t)$ in that episode to the probability distribution over the action space. In this paper, because we focus on decentralized control, we always use $s_t$ to denote the input of the policy model when it is clear from the context, instead of alternating to $o_t$ or $h_t$.

\subsection{Stochastic Game}
Stochastic games (also known as Markov games) \cite{shoham2008mas} generalize both Markov decision processes and repeated games. In repeated games, a given game (often in normal form) is played multiple times by the same set of players. With a little abuse of notation, a stochastic game is represented as a 6-tuple $\langle S, N, A, P, r, \gamma \rangle$ where $S$ is the state space and $N$ is a finite set of players. $A=A_1\times \cdots \times A_N$ is the joint action space where $A_i$ denotes the set of actions for player $i$. $P:S\times A\times S\rightarrow [0,1]$ is the state transition function. $P(s_{t+1}|s_t,{a_t})$ is the probability of the transition from state $s_t$ to state $s_{t+1}$ after joint action $a_t$. The payoff function $r=(r^1,\cdots,r^N)$, where $r^i:S\times A\rightarrow \mathbb{R}$, is real-valued for player $i$. A history $h_t$ of the game is a finite sequence $h_t = (s_1, a_1, \cdots, s_t)\in H$. Superscript is used to denote which player is, a player's \textit{behavior strategy} (the agent's policy) is defined as $\pi^i:H\times A_i\rightarrow [0,1]^{|A_i|}$, and their strategy profile (joint policy) is $\pi=\pi^1\times\cdots\times\pi^N$. Denoting all of the players expect player $i$ by ${-i}$, the strategy profile can be simplified to $\pi=\pi^i\times \pi^{-i}$. A \textit{pure strategy} of a player $i$ assigns all probability on a single actoin for each history $h_t$, i.e., $\pi^i:H\rightarrow A$. A \textit{mixed strategy} of the player is a probability distribution over all possible pure strategies. By Kuhn’s theorem, it can be shown that a mixed strategy is always equivalent to a behavior strategy and vice versa, given \textit{perfect recall}, which means that every player will not forget what they have done or observed in one episode. It is clear that perfect recall always holds if players make decesions in a history $h_t$. A Markov strategy is a behavior strategy with Markov, i.e., property $\pi^i(h_t, a_t)=\pi^i(s_t,a_t)$. Analogously, $\gamma$ is the discount factor.

The expected cumulative reward for player $i$ is formulated as $V^{\pi, i}(s)=\mathbb{E}^{\pi}[\sum_{t=0}^\infty \gamma^t r^i(s_t, a_t)|s_0=s]$ where $\pi=(\pi^i, \pi^{-i})$. A \textit{best response} $\pi^{i, \star}\in BR(\pi^{-i})=\argmax_{\pi^{i}} V^{\pi, i}(s)$, is a policy that achieves the highest expected cumulative payoff when $\pi^{-i}$ keeps fixed. Nash equilibrium is a strategy profile $\pi=(\pi^i, \pi^{-i})$ where $\forall i, \pi^i\in BR(\pi^{-i})$, so that no player can improve their payoff by deviating from it unilaterally. If $\forall i, V^{i,(BR(\pi^{-i}),\pi^{-i})} - V^{i,\pi}\le\delta$, $\pi$ yields a $\delta$-Nash equilibrium. The discounted-reward case is the less problematic: a Nash equilibrium exists in every stochastic game, as shown by folk theorem.

In two-player zero-sum stochastic games, $r^1(s_t,a_t)+r^2(s_t,a_t) = 0$ always holds true, and the expected cumulative reward of the opponent is $V^{\pi, -i}(s)=-V^{\pi, i}(s)$. It can be obtained from the definition that  $\max_{\pi^i}\min_{\pi^{-i}}V^{i}(s)\le \min_{\pi^{-i}}\max_{\pi^i} V^{i}(s)$. The underlying property of a maximin strategy is straight-forward: it guarantees the worst-case payoff when faced with an adversarial player. By minimax theorem, in two-player zero-sum games, the maximin value is equal to the minimax value and is called the \textit{value of the game}. It can be shown that every Nash equilibrium achieves the same unique value $V$ in two-player zero-sum games, and the Nash equilibrium is a maximin strategy.

In the remainder of this paper, we do not distinguish between the \textit{player} and the \textit{agent}, which are essentially the same, nor do the \textit{strategy} and the \textit{policy} or the \textit{payoff} and the \textit{reward}.

\begin{figure*}[!ht]
	\centering 
	\subfigure[Max Player's Policy model]{
		\label{Fig.sub.1}
		\adjincludegraphics[width=0.48\textwidth, trim={0 {0.3\width} {.1\width} {.05\width}}, clip]{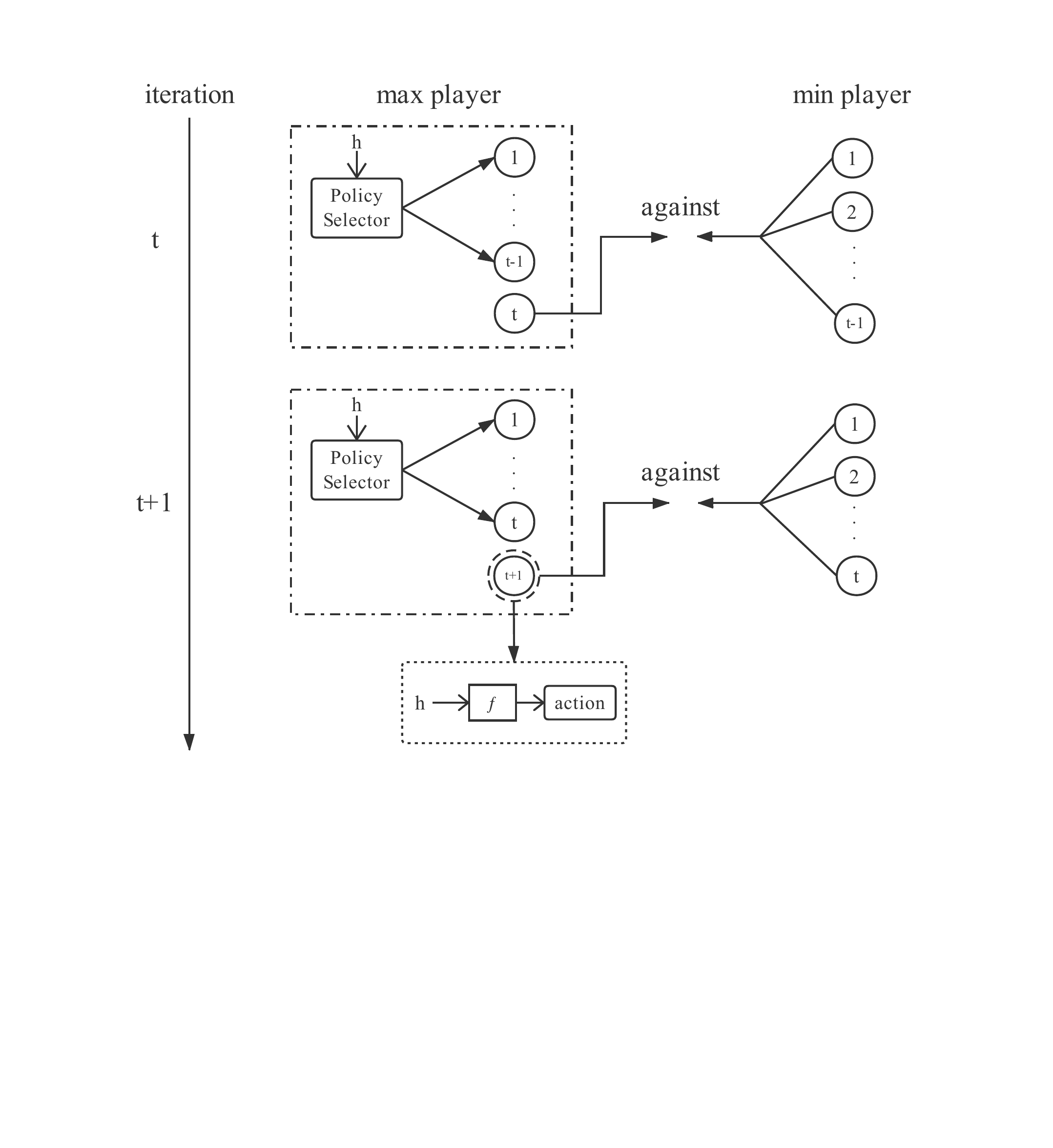}}
	\subfigure[Opponent's Policy model]{
		\label{Fig.sub.2}
		\adjincludegraphics[width=0.48\textwidth, trim={0 {0.3\width} {.1\width} {.05\width}}, clip]{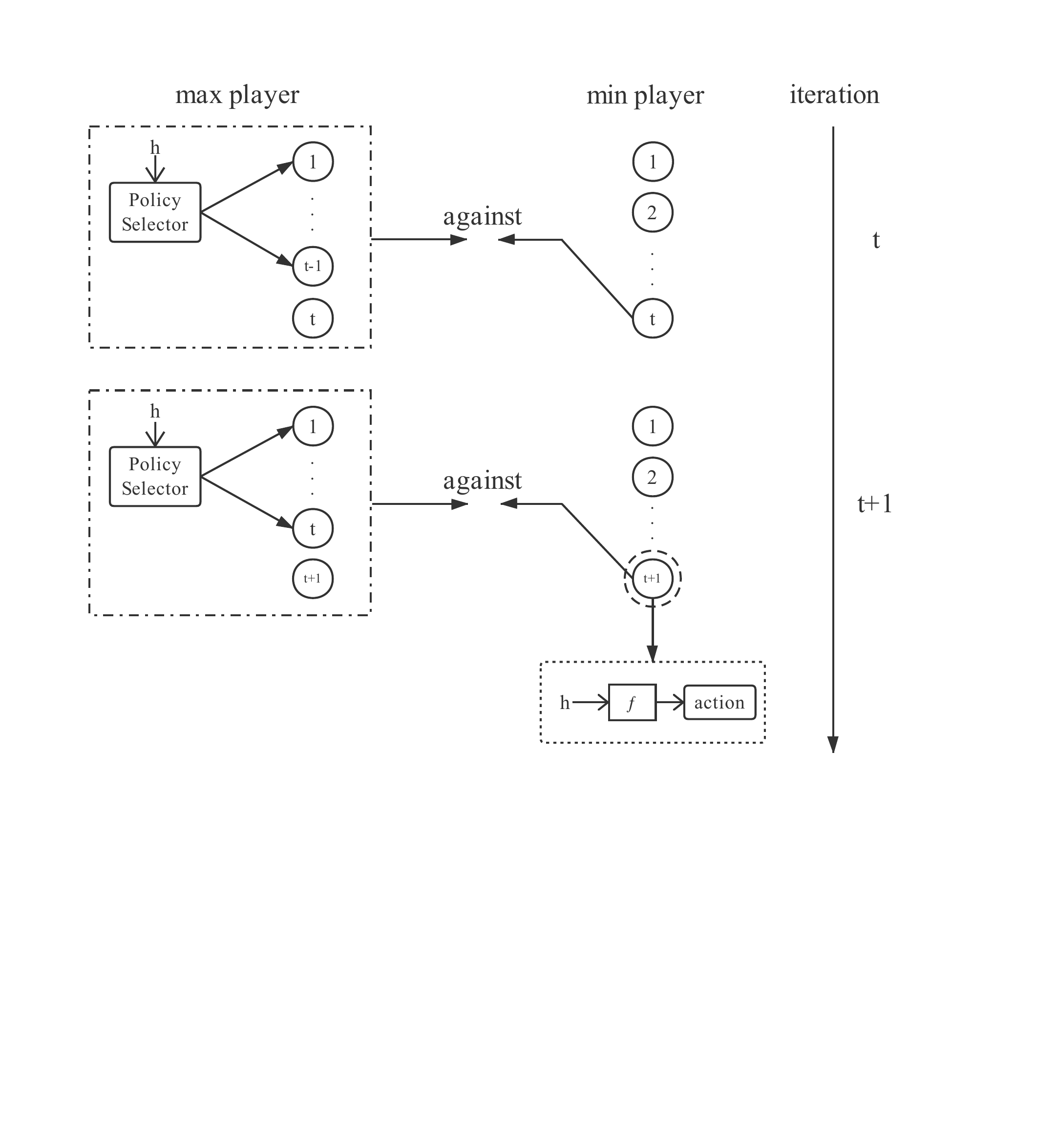}}
	\caption{Two policy generating approaches for the max player (player 1) and the min player (player 2). Each circle is a base policy which receives a history $h$ as its input and so does the policy selector $W$. As (a) shows, the max player's policy consists of two components: the policy selector and a base policy set. In iteration $t$, a new base policy is trained against the opponent pool and added to base policy set. Every iteration, the opponent trains a single policy $\pi_t^{-i}$ that tries to beat the FPEM model and is also added to the opponent pool after training. } 
	\label{Fig.1}
\end{figure*}

\section{Fictitious Play with Expanding Models}
In two-player zero-sum games, the decentralized fashion of solving for maximum expected utility is equivalent to solving the maximin strategy of the player, since the maximum for the opponent is the minimum for the player. Thus, the objective function can be represented as
\begin{equation}
	\pi^{i,\star}=\argmax_{\pi^{i}} \min_{\pi^{-i}} V^{\pi, i}(s).
\end{equation}

This problem can be solved by alternative optimization approaches. However, in complex environments, it is impossible to derive the closed-form solution for the max or min operator, albeit $V(s)$ is well formulated. It should be noticed that all the other players' strategies remain fixed during the max or min step when alternative optimization is used. Thus, the max or min optimization problem is reduced to a standard single-agent RL problem with new transition function composed of the environment dynamics and the other players' policy. We use neural networks as function approximators to represent the players' policies and value functions. By denoting player $i$'s and $-i$'s parameterized policies as $\pi^i_{\theta}, \pi^{-i}_\theta$ respectively, the new objectives can be represented as
\begin{equation}
	\theta^{i,\star}=\argmax_{\theta^{i}}\min_{\theta^{-i}} V^{(\pi^{i}_\theta, \pi^{-i}_\theta), i}(s).
\end{equation}
\begin{equation}
	\theta^{-i,\star}=\argmin_{\theta^{-i}}\max_{\theta^i} V^{(\pi^{i}_\theta, \pi^{-i}_\theta), i}(s).
\end{equation}

We assume that the opponent is a malicious adversary that always tries to minimize the expected return $V^{\pi}(s)$ to approximate the maximin process. Without loss of generality, we also assume that player $1$ is the max player, and player $2$ is the min player.

\subsection{Alternative Optimization with Single Model }
The players' policies are both represented by neural networks so that the extremum operator can be approximately solved with deep reinforcement learning techniques, and alternative optimization can be readily applied. In the max or min step, the corresponding player tries to solve for the optimal policy when the opponent's policy is fixed by updating the parameters. Since a neural network is used as the function approximator and the extremum operation is coupled with the opponent's policy, the underlying problem is \textit{instability}. As the training process continues, the current trained policy is no longer a best response or even loses when playing against some opponents that used to be defeated, namely, forgetting old opponents. So the training process may fluctuate or even diverge, especially in cases where the opponent's policy is crucial. The instability here is inherent in the policy model due to back-propagation updating. 

In independent RL, because the unawareness of the other players and simultaneous learning, the dynamics of the environment is much more complex and induces the instability, irrespective of what policy model is used.

\subsection{Opponent Pool with Single Model}
To stabilize the training process, i.e., to guarantee the performance when facing defeated opponents, instead of competing with a single policy model, the opponent's historical policies are kept in a buffer, which is known as the \textit{opponent pool}. AlphaGo \cite{silver2016alphago} also used an opponent pool to stabilize training by preventing overfitting to the current policy being trained. It added the current policy to the opponent pool every 500 iterations and samples a policy to compete within each iteration (a mini-batch of $n$ games). 

In this paper, we add an opponent policy to the pool after every min iteration and uniformly sample a policy for each game from the pool. The max player trains the current policy to maximize the expected return against the mixture of policies from the opponent pool. The opponent's policy $\pi^{-i}_t$ that is added to the pool is a best response to the max player's policy $\pi^i_t$, which tracks the process of Brown's original fictitious play. Because all of the opponent's policies are kept and form a mixture model of policies, this helps to alleviate the problem of forgetting so that the performance will not have a sudden drop when facing a defeated opponent. For simplicity and to adhere to fictitious play, we just uniformly sample from the opponent pool. However, some other distributions can be used in practice, e.g., a distribution that is proportional to the ranking of the opponent's policies \cite{OpenAI_dota,alphastar_nature}. The opponent pool can be treated as a mixed strategy, as it selects each policy (albeit not a deterministic policy) from some distribution before the game and commits to it the whole game.

\begin{algorithm}[!ht]
\caption{FPEM}
\label{alg:algorithm1}
\textbf{Input}: number of maximum base policies $T$, base policy set $BP$, opponent pool $OP$, a reservoir memory $M$, policy selector network $W$ and target network $W^\prime$\\
\textbf{Output}: policy selector $W$ and corresponding base policies
	\begin{algorithmic}[1] 
		\For  {iteration $t$ = $1:T$}
		\For {$i\in\text{players} [N]$}
		\State initialize a new $\pi^i_t$
		\For {$e$ in $n$ episodes}
		\If {$i$ is max player}
		\State $\pi^{-i}\sim Unif(OP)$ (random if empty)
		\Else
		\State  $$\pi^{-i}=\left\{
		\begin{aligned}
		\pi^{-i}_t &   & w.p. &&1/t \\
		W^\prime\circ \pi^{-i}_{1:t-1}   &   & w.p. && 1- 1/t \\
		\end{aligned}
		\right.
		$$
		\EndIf
		\State play the game and collect trajectories
		\If {$i$ is min player and $t\ge 2$}
		\State store $(h,j)$ in $M$ ($j$ is the index of the policy executed at $h$)
		\EndIf
		\If {$e==0$ mod (learning\_frequency)}
		\State update $\omega_t$ with SGD on loss $$\mathbb{E}_{(h,j)\sim M}[-\log W(h,j|\omega_t)]$$
		\State optimize $\pi^i_t$ with collected transitions by RL algorithm
		\EndIf
		\EndFor 
		\State $BP \gets BP\cup \pi^i_t$ (or $OP \gets OP\cup \pi^i_t$)
		\EndFor 
		\State set $W^\prime=W$
		\EndFor 
	\end{algorithmic}
\end{algorithm}

\subsection{Fictitious Play with Expanding Models}
Instead of constantly updating a single policy model in the max step, we use an expanding model approach that continually generates new policies and combines all of them. The new policy model is named \textit{fictitious play with expanding models} (FPEM), and consists of a base policy set $BP$ and a policy selector $W: H\rightarrow [0,1]^{|BP|}.$ FPEM is initialized with an empty base policy set, and the selector $W$ has no options. After every max step, a newly trained base policy $\pi_t^i$ is added to $BP$, and thus $W$ has a new optional policy. Only the parameters of current training base policy $\pi_t^i$ and $W$ are updated, and the previously trained base policies keep fixed (if there are). During the min player's step, the max player sample a policy from the base policy set follow a specified distribution (e.g., uniform distribution) at the begining of each game, and the experience pair $(h, j)$ is stored in memory $M$, where $h$ is the history and $j$ is the index of base policy executed at that history. $W$ then learns from the stored experiences. To learn the policy selector with limited experiences, reservoir sampling \cite{1985reservoir} is used to store these experiences from those best response policies. Let $W$ be a neural network and parameterized by $\omega$; using $W\circ \pi^i_{1:t}$ denotes the FPEM model, i.e., the combined policy selector and base policy set $\{\pi^i_1, \pi^i_2, \cdots, \pi^i_t\}$. Thus the objective of FPEM is 
\begin{equation}
\omega_t=\argmin_{\omega} \mathbb{E}_{(h,j)\sim M}[-\log W(h,j|\omega)]\label{Wobj}
\end{equation}
\begin{equation}
\theta^i_t=\argmax_{\theta^i} V^{(\pi^{i}_{t}, \pi^{-i}), i}(h),\label{BPobj}
\end{equation}
, where $\theta^i_t$ represents the parameters of the base policy $\pi_t^i$. Since we aim at solving for the max player's policy, thus we use FPEM as the max player's policy model and intentionally use the opponent pool for the min player for simplicity. In fact, both players can adopt FPEM as their policy model. Figure \ref{Fig.1} shows both the process of policy generating in FPEM and the opponent's policy generating with the opponent pool.

We make policy selector $W$ a behavior strategy that selects a base policy at every action history, and mix these base policies in the behavior strategy level. $W$ and $\pi_t^i$ are trained while parameters of $\pi^i_{1:t-1}$ are not updated, which is a trade-off between the stability and adaptability. Keeping and incorporating previously trained base policies will overcome the catastrophic forgetting problem, which helps stabilize training. Besides, adding a new base policy and retraining the policy selector $W$ will improve the adaptability of FPEM since the new base policy serves as a sub-model to make FPEM adapt to the newly updated opponent pool.

The learning of average policy in FSP or NFSP is a policy distillation process, where the average policy constantly learns from the inputs and outputs of best response models. FPEM explicitly expands with these best response models and form a hierarchical structure, which potentially improves learning efficiency. Another potential advantage is that using a behavior strategy rather than a mixed strategy can improve the robustness when evaluation with finite simulations. Although there is an equivalence relationship between a mixed strategy and a corresponding behavior strategy given perfect recall, they will receive the same cumulative reward only in expectation. A single base policy is trained against a specified opponent, so it may not be that reasonable in all states. However, FPEM integrates all these base policies and makes decisions in every history so that multiple policies work jointly during one episode.

The process of FPEM is summarized in Algorithm \ref{alg:algorithm1}. The max player (take player 1 as max player) trains a best response $\pi^1_1$ to the opponent and adds it to base policy set $BP$. The min player trains a best response $\pi^2_1$ against $\pi^1_1$. In iteration 2, the max player trains $\pi^1_2$ and adds it to BP; the min player trains $\pi^2_2$ against $\pi^1_1,\pi^1_2$ with equal probability and $W$ learns from $(h, j)$ pairs, where $j$ is the index of the base policy executed at the history state $h$. At the end of this iteration, the target network $W^\prime$ is replaced by $W$ so that in the next iteration, $W^\prime$ combined with $\pi^1_1,\pi^1_2$ (denoted by $W^\prime\circ \pi^{1}_{1:2}$) is a uniform mixture of trained policies. In iteration 3, the min player trains against $W^\prime\circ \pi^{1}_{1:2}$ and $\pi^1_3$ with probabilities $2/3,1/3$, respectively, which is equivalent to competing with $\pi^1_1,\pi^1_2,\pi^1_3$ with equal probabilities. New $(h, j)$ pairs are stored in the reservoir memory and $W$ retrains. Since each $W$ is an approximation of the specified distribution (e.g., uniform distribution) over $\pi^1_{1:t-1}$ and retrains to incorporate the newly trained $\pi^1_t$, FPEM maintains the specified mixture in behavior form.

Notice that in \ref{alg:algorithm1}, the uniform distribution version of FPEM is described, but FPEM is not restricted to uniform distribution. In fact, the policy selector can learn from any proper distribution. In this paper, we mainly focus on uniform distribution.

The latest base policy of both players always learns a best response to the opponent pool, so the FPEM potentially enjoys the convergence property in two-player zero-sum games, as it tracks the process of fictitious play.  The above implies that if FPEM converges in a two-player zero-sum game, then it converges to a Nash equilibrium, which is also a maximin strategy. 

In the experiments, we also evaluate a non-alternating learning variant of FPEM (FPEMv1), where each player updates his policy model simultaneously. Learning a new best response and playing with a combination of other models and the current learning model, the simultaneous learning process leads to a off-policy problem. So for the simultaneous learning version, we use off-policy RL algorithms or apply a clipped importance sampling weight on the critic (value funtion).

\label{grid_world}
\begin{figure*}[h]
	\centering 
	\subfigure[Map1]{
		\label{Fig2.sub.1}
		\adjincludegraphics[width=0.3\textwidth, trim={0 {0.7\width} {.3\width} {.05\width}}, clip]{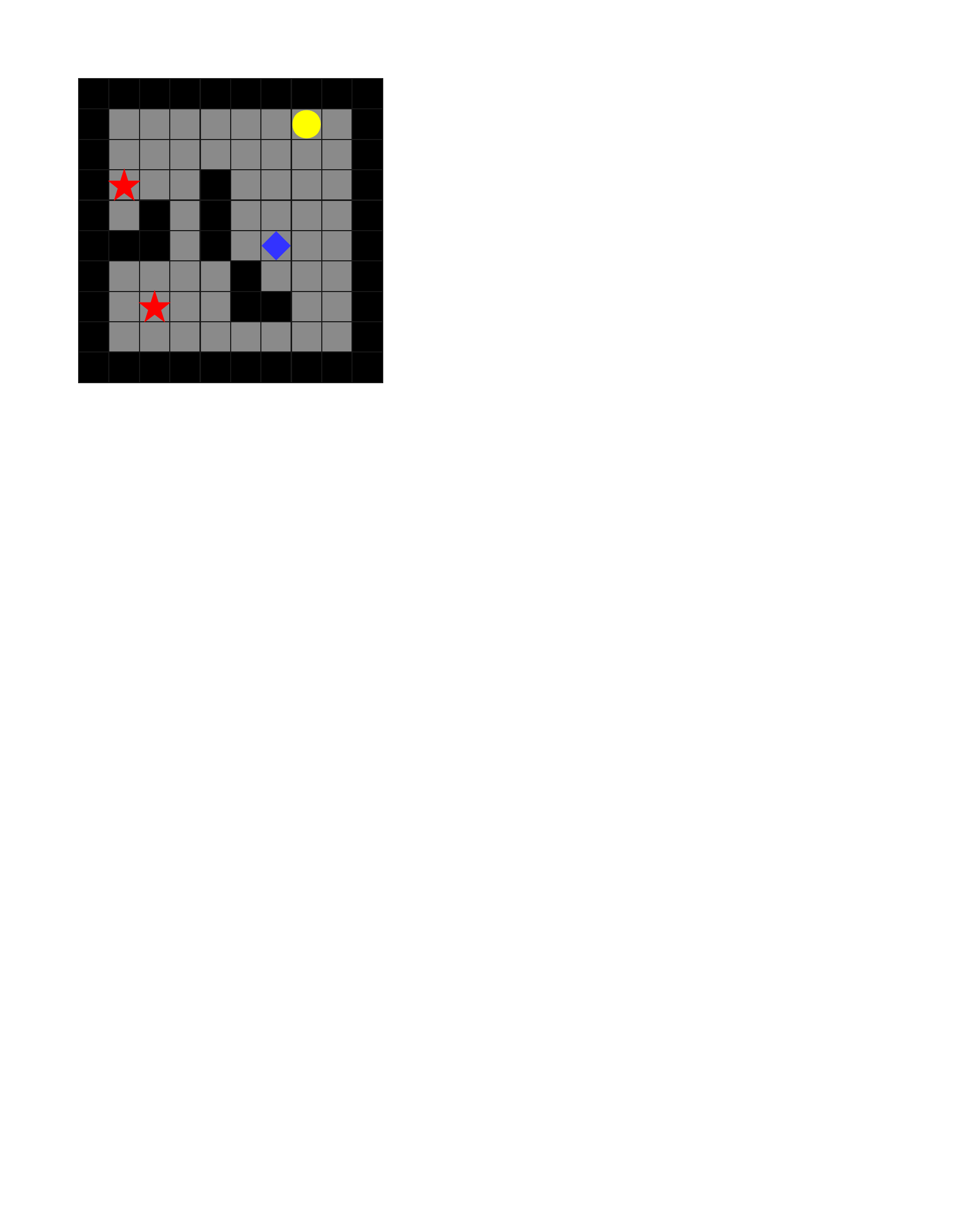}}
	\subfigure[Map2]{
		\label{Fig2.sub.2}
		\adjincludegraphics[width=0.3\textwidth, trim={0 {0.7\width} {.3\width} {.05\width}}, clip]{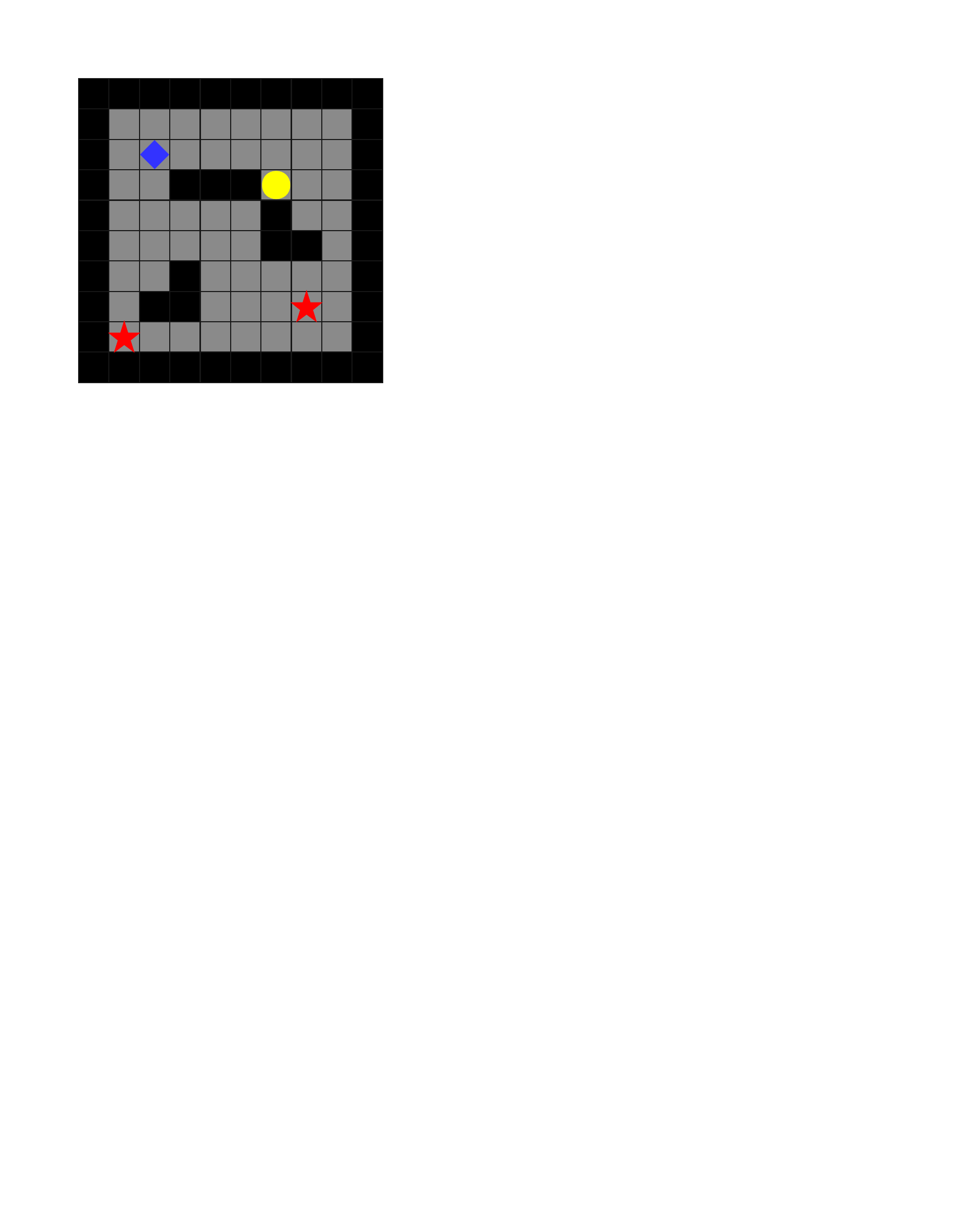}}
	\subfigure[Field of View] {
		\label{Fig2.sub.3}
		\adjincludegraphics[width=0.23\textwidth, trim={{0.0\width} {0.875\width} {.575\width} {.0\width}}, clip]{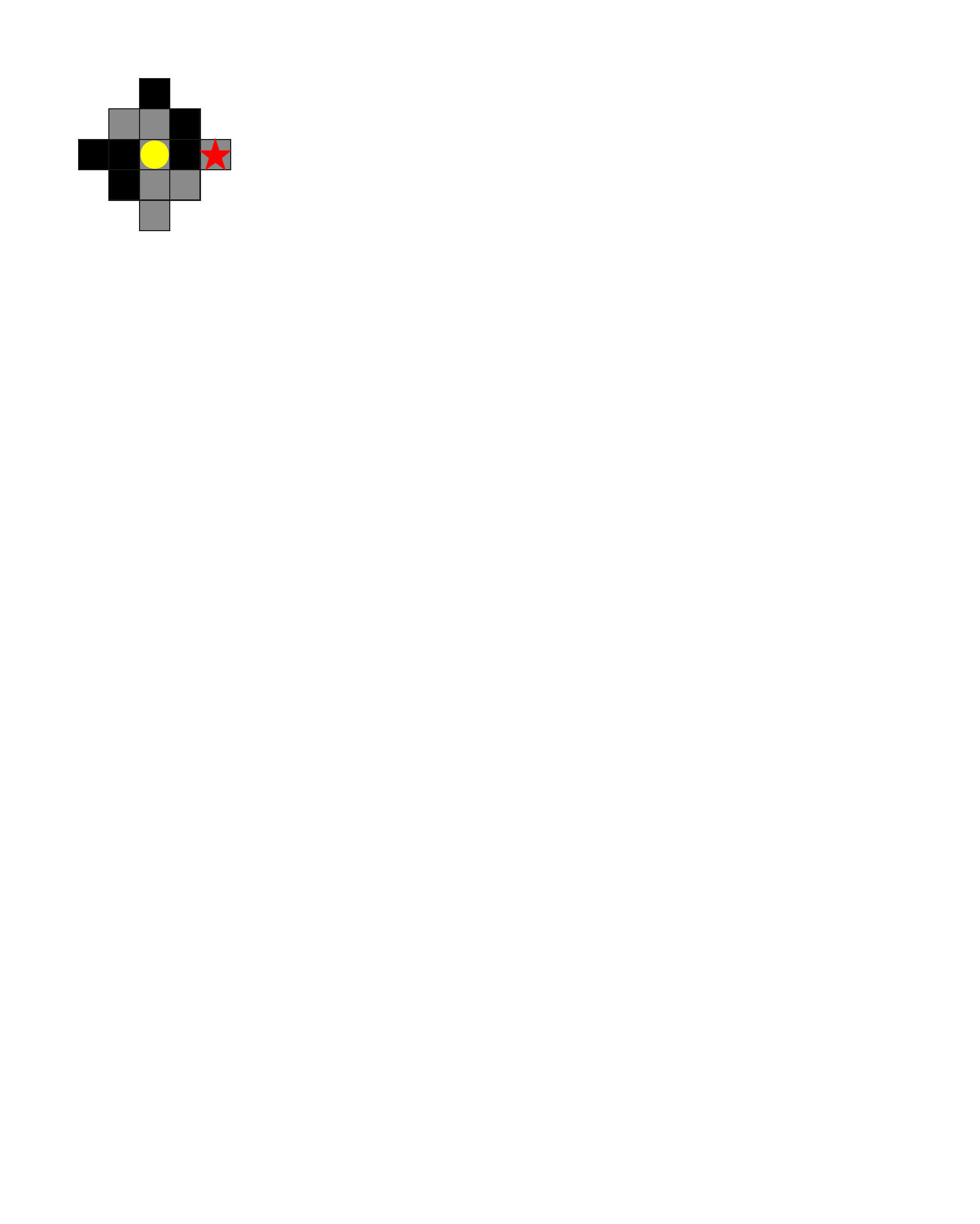}
	}
	\caption{Visualization of two initialized Treasure Hunting maps and the field of view of a player. In (a) and (b), players are represented with a diamond (max player) and a circle (min player). The movable areas are denoted by hollow grey squares, and solid black squares are walls that block players. Treasures are denoted by red stars and randomly spawned in an empty grid. (c) shows the field of view of one player. It is a diamond-shaped area with a maximum of 13 grids, and player locates in the central grid. This area shrinks when the player is near the boundary. } 
	\label{Fig2}
\end{figure*}

\section{Experiments}
 The proposed FPEM is first evaluated in Kuhn poker as a case study and then two stochastic games. Kuhn poker is an extensive-form imperfect-information game. The two stochastic games are a two-player zero-sum treasure hunting environments and Mini-RTS from ELF platform \cite{minirts}. We also note that the last two games are symmetric, i.e., for any permutation of players, the policy space and the reward for that player will not change, while Kuhn is not symmetric. So for symmetric games, we only use FPEM as the max player's policy model and maintain a pool for the opponent, and after training, only FPEM is evaluated. For unsymmetric games, FPEM models are used for all the players. 
 
 \subsection{Case Study: On Kuhn Poker}
 Two-player Kuhn poker is a toy poker game with 3 totally-ordered cards, where each player starts with a private card and assigned 2 chips for one game and antes 1 chips to play. Players can choose to  bet, check or fold and the betting is limited to 1. The game ends with a showdown or one player folds.
 
 $\text{NashConv}(\pi)=\sum_{i}^N \max_{pi^{i,\prime}} V^i(\pi^{i,\prime}, \pi^{-i})-V^i(\pi^{i}, \pi^{-i})$ is commonly used in poker AI, which measures how much players gain by unilaterally switching to a best response. It also measures the distance from a Nash equilibrium, where a NashConv value of $\delta$ yields a $\delta$-Nash equilbrium. This measure can easily be obtained in Kuhn poker. The reason for choosing Kuhn poker because there is a transition dynamics yet simple enough to get a close-form measure rather than caring for the poker game.
 
 To verify the expanding models, we compare FPEM with original NFSP and based on the open-source library OpenSpiel \cite{OpenSpiel}. In FPEM, each expanded model is a deep Q-nets (DQN). After some episodes of best response learning, this policy stops learning, and a new base policy begins to learn. Once the second base policy begins to learn (the second outer iteration), the policy selector $W$ learns the mixture at each state, which is equivalent to the uniform distribution over all the trained base policies. Because DQN is an off-policy RL algorithm, the experiences from previous trained models can be put into the replay buffer. Thus the implementation of FPEM mainly differs from NFSP in that the average strategy is expanding with newly trained models. Since the evaluation with NashConv requires a behavior strategy rather than a pool of trained best response, RSPO is not compared currently.
  \begin{figure}[h]
  	\centering
  	\adjincludegraphics[width=0.7\textwidth, trim={0 {0.0\width} {.0\width} {.0\width}}, clip]{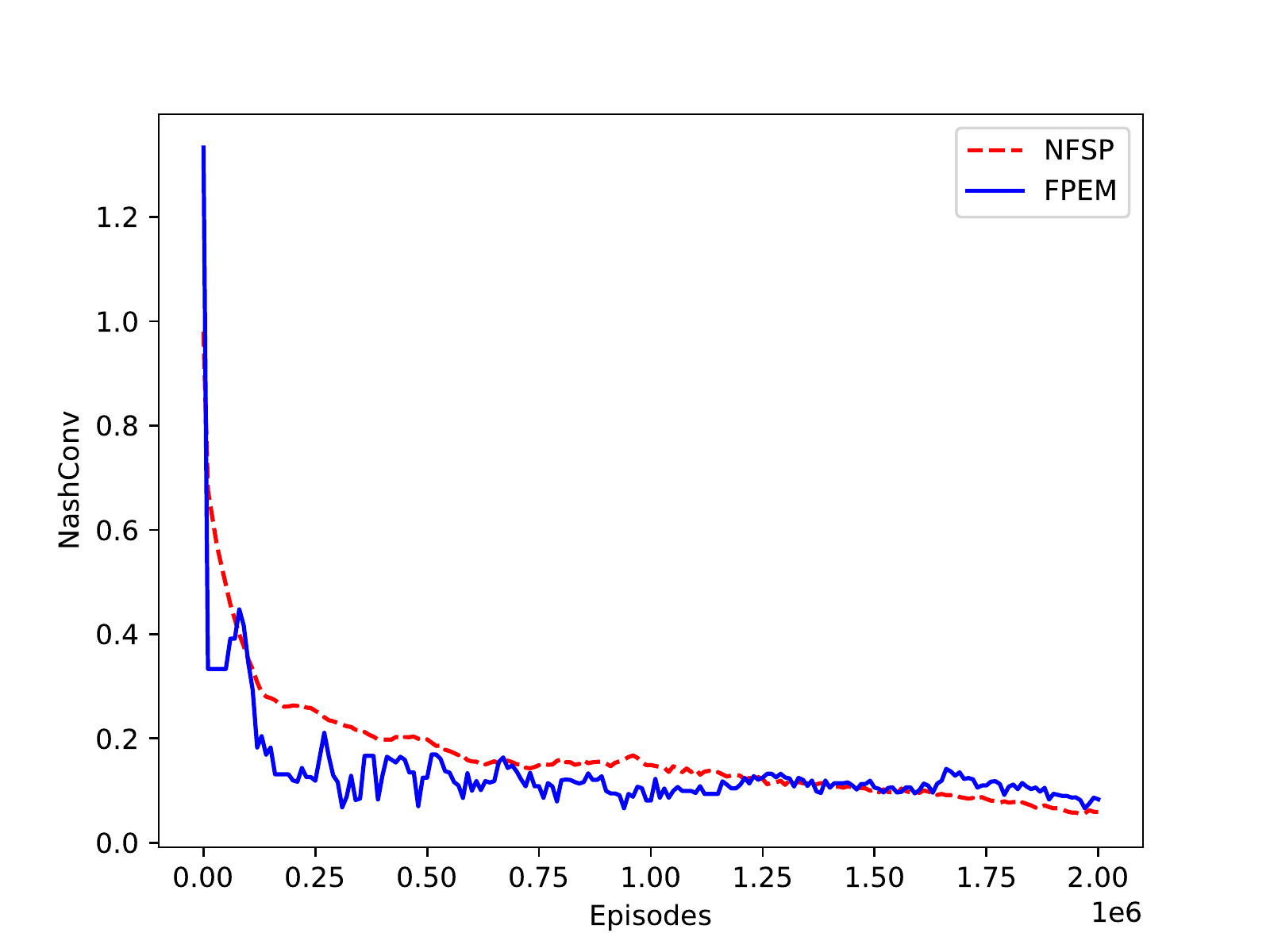}
  	\caption{NashConv in 2-player Kuhn poker.}
  	\label{Fig.leduc}
  \end{figure}
 
 For NFSP, the anticipator $\eta=0.1$ and the two policy networks are both 1-layer MLP with 64 units, which are the same as the original paper (varying hidden units seems to make little difference). For FPEM, the policy selector $W$ is also a 1-layer MLP with 64 hidden units. During the learning process of $W$, the target selector network $W^\prime$ is also used to maintain the distribution over $t-1$ base policies, and current learning base policy is executed with probability $\frac{1}{t}$. So, at the begining of one game, with probability $1-\frac{1}{t}$, the target selector $W^\prime$ with $t-1$ base policies is chosen to follow for a game, and the new base policy is chosen with probability $\frac{1}{t}$. $W^\prime$ is update by $W$ at the end of an iteration. The number of maximum base policies is set to 40. Each base policy learns for $5\times 10^4$ episodes.

Figure \ref{Fig.leduc} shows that with expanding models, the average policy is faster to achieve a lower NashConv. In the first 0.05M episodes, the NashConv is directly from the $\epsilon$-greedy$(Q)$ policies. After that, the NashConv is calculated from the training average policy (multiple $\epsilon$-greedy$(Q)$ in behavior level), in corresponding with NFSP. After a long-term training, the NashConv of both algorithm reachs to a very low value. Because adding too many sub-models requires more samples and complex network architectures for the policy selector to learn, we do not run FPEM with more than 50 base policies. With multiple runs, the final NashConv of NFSP in Kuhn poker seems to plateau at $0.03\pm 0.05$ after 10M episodes, while FPEM only reaches $0.08\pm 0.03$ with maximum of 5M episodes, thus FPEM learns faster than NFSP with limited samples. Possibly due to the stochastic nature of poker, and DQN is a deterministic sub-model, it requires more samples or hyper-parameter searchs for FPEM to achieve a better performance in the long run.

\subsection{On Stochastic Games}
In the latter two environments, due to the long horizon and sparse reward properties, we use policy-gradient based RL approaches and stack the last $K$ frames to approximate history $h_t$. In treasure hunting, PPO \cite{schulman2017ppo} is the solver for best response and A3C \cite{a3c} for Mini-RTS. However, FPEM is not coupled with PPO, A3C, and any other efficient solvers can be used. As reported in \cite{nfsp_rts19} and the original ELF paper, A3C implemented by the ELF platform slightly outperforms PPO against two built-in AIs.
\subsection{Environment Description}

\noindent\textbf{Treasure Hunting} \indent The treasure hunting game happens in an $8\times 8$ grid world. With randomly generated obstacles in the grid, each player is able to receive a partial observation of the environment, which is a diamond-shaped area. Fig \ref{Fig2} two randomly initialized map and the observable areas. The available actions for each player are \{UP, DOWN, LEFT, RIGHT\}. A player will be rewarded 1 if he obtains a treasure, and the opponent receives a reward -1 at the same time. If the player who has owned a treasure gains a second one, he receives a reward 2, and -2 for the opponent. When two players gain a treasure simultaneously, both will receive a reward 0, and this treasure disappears in the map. A grab occurs when one player who carries a treasure enters the same grid with the opponent, then the treasure loses to the opponent and the reward will be -2. An episode usually lasts for 60-80 ticks and with a maximum of 100 ticks. Each player will only receive the reward signal when the player gains or loses a treasure.

\noindent\textbf{Mini-RTS} \indent Mini-RTS is a partially observable two-player zero-sum game environment on the ELF platform. It is a miniature custom-made RTS game that captures all the underlying dynamics of StarCraft (fog-of-war, resource gathering, troop building, defence/attack with troops, etc.). The observation is composed of spatially structured (20-by-20) abstractions of the current game environment with 22 channels. Although Mini-RTS offers micro-commands, the actions used in the experiments are nine strategic hard-coded global commands. Every game terminates at a maximum of 30000 ticks (an average game lasts for 1000-6000 ticks). There are two rule-based built-AIs named SIMPLE and Hit-aNd-Run (HNR) respectively (a human player has a win rate of $90\%$ and $50\%$ against SIMPLE and HNR in 20 games).

\begin{table*} [ht]
	\centering 
	\caption{Winning rate against adversary in Mini-RTS.} 
	\scalebox{.72}{
		\begin{tabular}{lcccccccccc}  
			\toprule
			Model & OP1 (\%)  & OP2 (\%)& OP3 (\%) & OP4 (\%) & OP5  (\%)& OP6  (\%)& OP7 (\%)& OP8 (\%) & OP9  (\%)& OP10 (\%)  \\  
			\midrule
			FPEM      & 49.6$\pm$1.6  & 49.9$\pm$1.3 & 50.2$\pm$1.4 & 51.0$\pm$1.0 & 50.8$\pm$1.1 & 50.4$\pm$1.4 & \bf{50.7$\pm$1.2} & 50.7$\pm$1.8 & \bf{51.0$\pm$0.2} & 50.5$\pm$1.2 \\  
			FPEMv1  & \bf{50.3$\pm$1.7} & \bf{51.2$\pm$1.4} & \bf{51.8$\pm$1.2} & \bf{52.6$\pm$1.1} & \bf{52.5$\pm$0.9} & \bf{50.8$\pm$1.3} & 50.2$\pm$1.6 & \bf{51.2$\pm$0.7} & 50.9$\pm$1.1 & 50.3$\pm$1.5  \\  
			SMv1      & 49.5$\pm$0.2 & 49.8$\pm$0.3 & 50.2$\pm$0.4 & 50.4$\pm$0.4 & 50.1$\pm$0.3 & 49.8$\pm$0.4 & 50.3$\pm$0.4 & 50.1$\pm$0.3 & 49.7$\pm$0.5 & \bf{50.5$\pm$0.4} \\
			SMv2      & 49.2$\pm$1.5 & 48.6$\pm$1.6 & 48.1$\pm$1.5 & 48.5$\pm$1.5 & 47.8$\pm$1.5 & 49.2$\pm$1.2 & 48.9$\pm$1.3 & 48.5$\pm$1.6 & 48.6$\pm$1.2 & 48.2$\pm$0.5 \\
			OPPO      & 49.6$\pm$1.2 & 50.6$\pm$1.3 & 49.8$\pm$1.5 & 50.2$\pm$1.4 & 50.4$\pm$1.3 & 49.9$\pm$1.4 & 50.1$\pm$1.3 & 49.9$\pm$1.1 & 50.2$\pm$1.1 & 49.7$\pm$1.4 \\
			\bottomrule
		\end{tabular}}
		\label{Tbl.rts}
	\end{table*}

\subsection{Experimental Settings}
In the treasure hunting experiment, the base policies of two players are both 4-layer MLP. The policy selector $W$ is a 4-layer MLP. For a single model policy, a 5-layer MLP is used. ReLu is the nonlinear activation function in all MLPs. The player who has the higher return wins in a game, and ties with the opponent only when both players' returns are 0. In each max or min step, it alternates to train a new base policy when 
\begin{equation}
\label{stopcri}
\frac{\{ \sharp \text{ of } win - \sharp \text{ of } lose\} }{\sharp \text{ of } {episodes}} > \delta_t, 
\end{equation}
or $100K$ episodes are reached. (\ref{stopcri}) is computed using the most recent $6K$ episodes. $\delta_t$ is set to 0.2.

In Mini-RTS, the base policy model is a simple CNN which is the same as \cite{minirts}, with two heads for the policy and the critic respectively. $W$ shares the CNN parameters with base policy and connected by one FC layer. Every base policy learns for $300K$ episodes. The number of maximum base policies is 10 in both environments.

\subsection{Results and Analysis}
For both experiments, we save the trained model after each iteration and evaluate these saved policy models. We evaulate on 3 seeds for both experiments. Unlike Kuhn poker, there is no such closed-form measure of NashConv, so the metric we use is how much a trained policy model loses against a trained malicious opponent, which measures how much a policy can be exploited. 

In treasure hunting, the FPEM, FPEMv1 are compared with five other approaches. Since we aim to solve for the policy of the max player, only the max player is evaluated. Note that both games are symmetric, the players are homogeneous, so the evaluation of the max player is enough. Besides, all of the policy models are initialized with pre-trained models via self-play, at a fraction of episodes used in one iteration to facilitate early-stage learning. SMv1 is the alternative optimization approach, where each player uses a single model. SMv2 is a variant of SMv1, where the min player is replaced by an opponent pool with a maximum of 10 base policies. Each base policy in the pool is a best response to the max player's corresponding policy. NFSP uses the same network structure as in SMv1, except there is another average policy, and both players learn simultaneously. The anticipatory parameter $\eta$ in NFSP is 0.25 which performs best among $\{0.1, 0.2, 0.25, 0.5\}$. In the OPPO, both players use the uniform opponent pool. For PSRO, each item of the expected utility tensor is computed via $10K$ simulations, and the meta-strategy is solved by regret matching \cite{hart2000simple}, as it is comparable to PRD solver in PSRO and straight-forward to implement. Besides, the same exploratory strategies for PSRO are used as in \cite{PSRO}.

\begin{figure}[!h]
	\centering
	\adjincludegraphics[width=0.7\textwidth, trim={0 {0.0\width} {.0\width} {.0\width}}, clip]{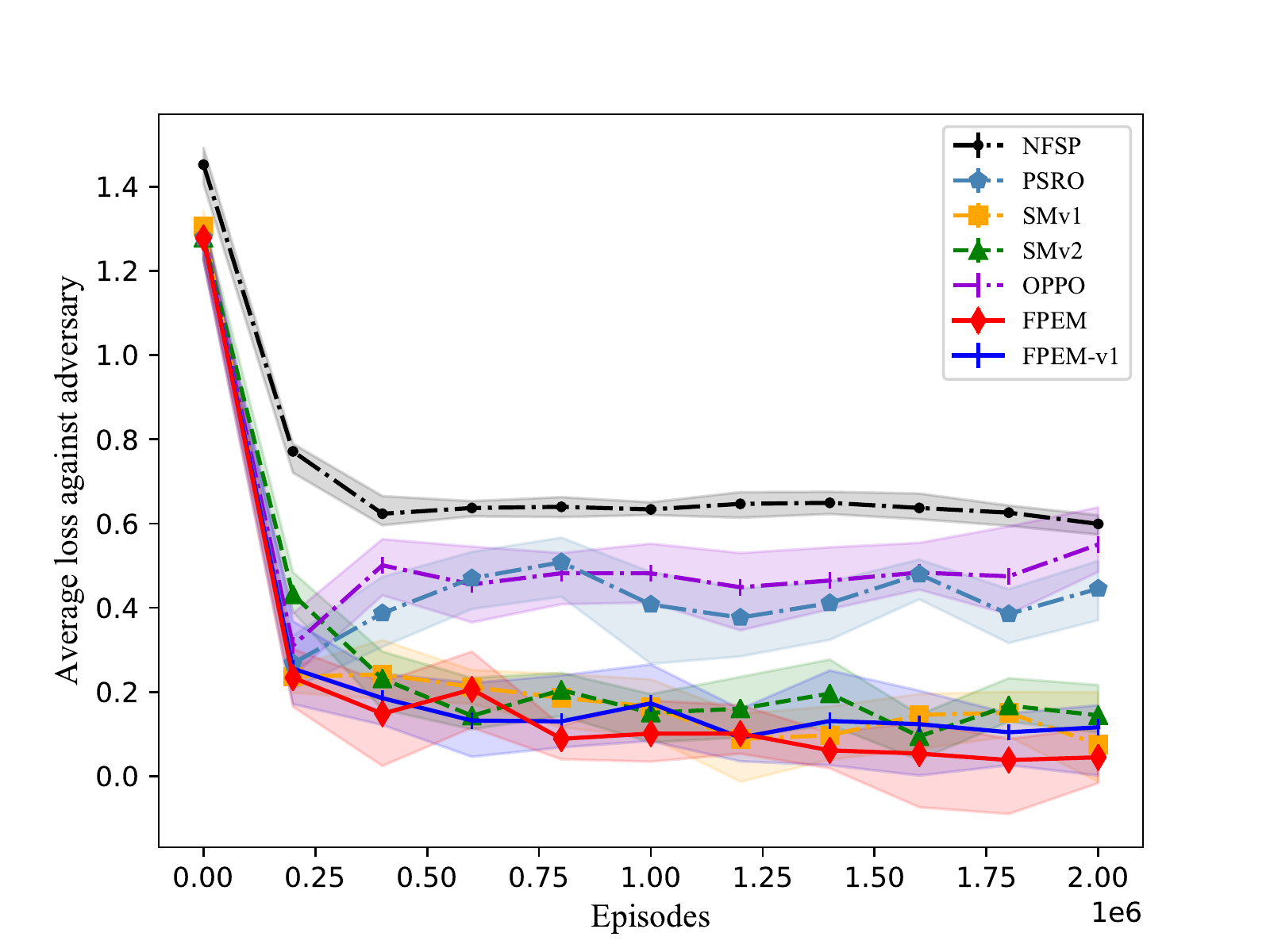}
	\caption{Average losses against malicious adversaries in Treasure Hunting. The loss is computed by playing against retrained adversaries, and each point is the average loss tested using $2\times 10^4$ episodes.}
	\label{Fig.4}
\end{figure}

For each model, we retrain 10 models with a 4-layer MLP to approximate the malicious adversaries, and these adversaries are trained until the expected return plateaus or a maximum of $300K$ episodes are reached. Figure \ref{Fig.4} demonstrates that SMv2, FPEM and FPEMv1 are less exploited when the opponent pool increases and the FPEM is slightly better than the other two. The shaded areas denote the average losses with the standard deviation from 3 runs. SMv1 fluctuates as training proceeds since it may forget how to respond to old opponents. And due to the deeper networs, SMv1 seems to comparable with SMv2. Although NFSP does not fluctuate and is with an extremely low variance, it appears to learn slowly. In the OPPO, due to sampling a base policy to follow for a whole game, the mixture of the pool is vulnerable. The overall performance of PSRO is better than the uniform version PSRO (OPPO), but may still be restricted by the accuracy of the expected utility tensor, where $10K$ simulations may be insufficient to get a good approximation. However, $10K$ simulations on each item are the trade-off between the time costs and the accuracy. Moreover, the average losses of FPEM decrease, thus the policy loses less and less against the malicious opponents and approximates the maximin strategy.

\begin{figure*}[!h]
	\centering 
	\subfigure[Simple-50]{
		\label{Fig5.sub.1}
		\adjincludegraphics[width=0.48\textwidth, trim={0 {0.0\width} {.0\width} {.0\width}}, clip]{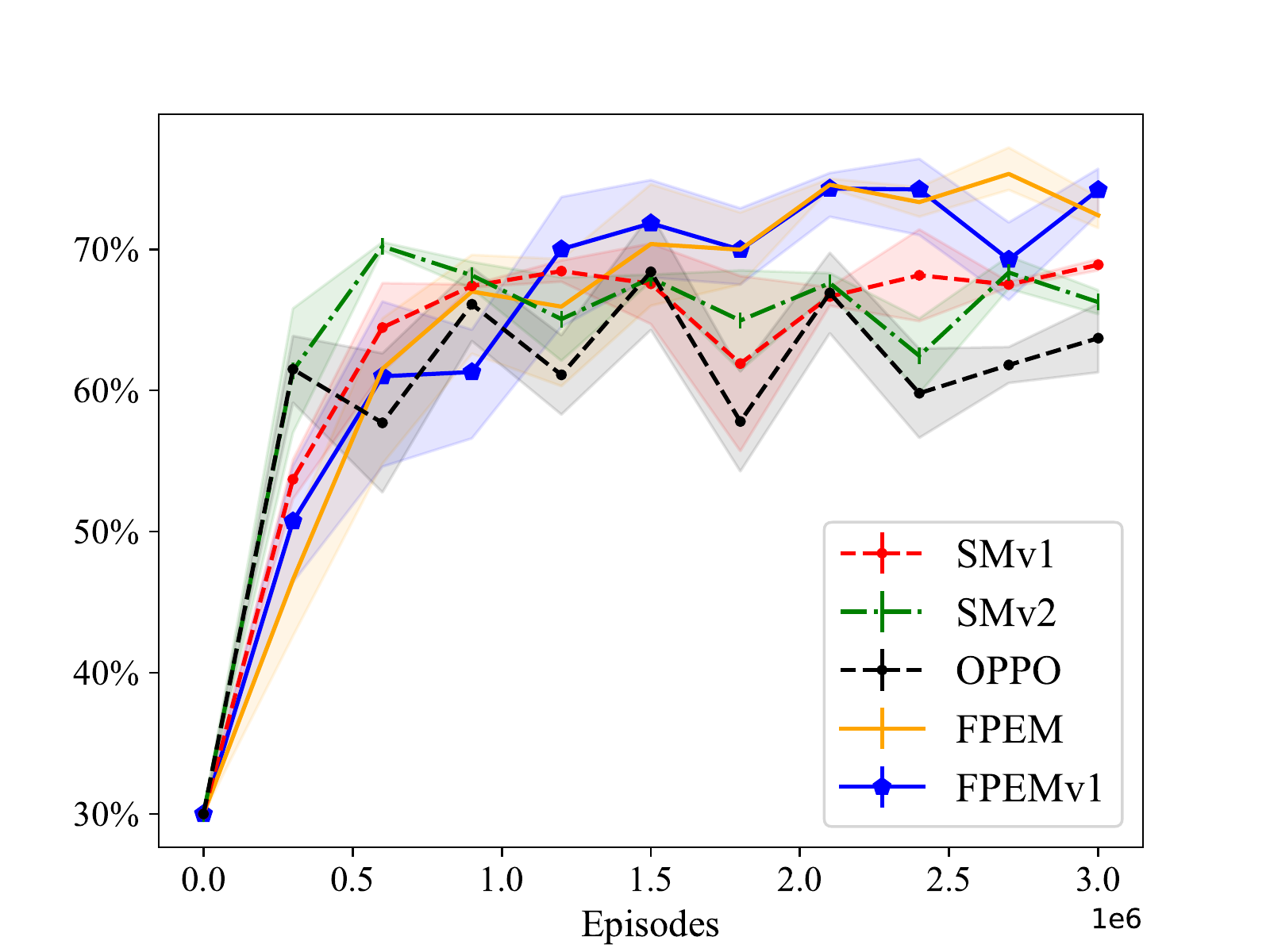}
	}
	\subfigure[Simple-10]{
		\label{Fig5.sub.2}
		\adjincludegraphics[width=0.48\textwidth, trim={0 {0.0\width} {.0\width} {.0\width}}, clip]{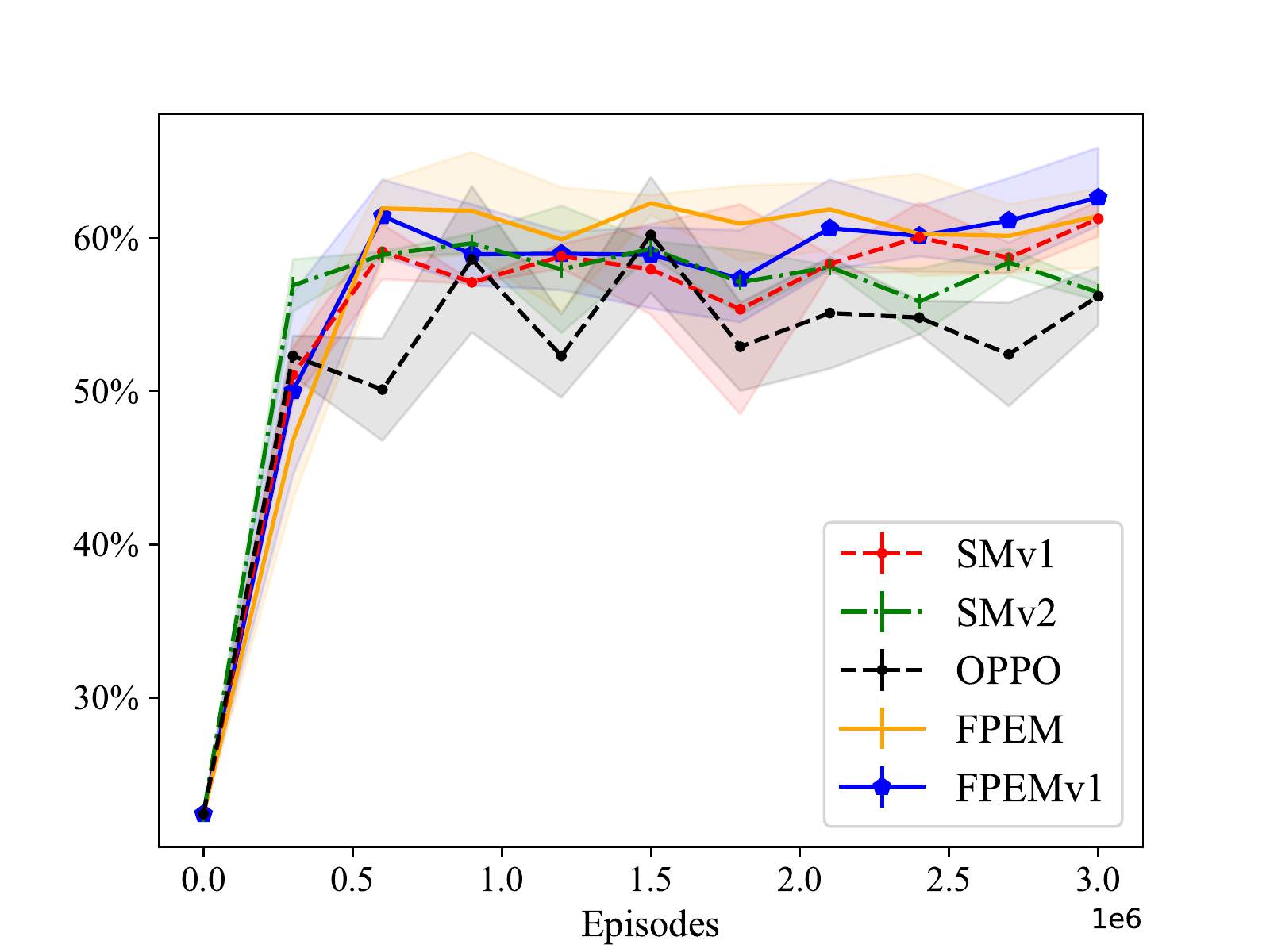}}
	\subfigure[HNR-50]{
		\label{Fig5.sub.3}
		\adjincludegraphics[width=0.48\textwidth, trim={0 {0.0\width} {.0\width} {.0\width}}, clip]{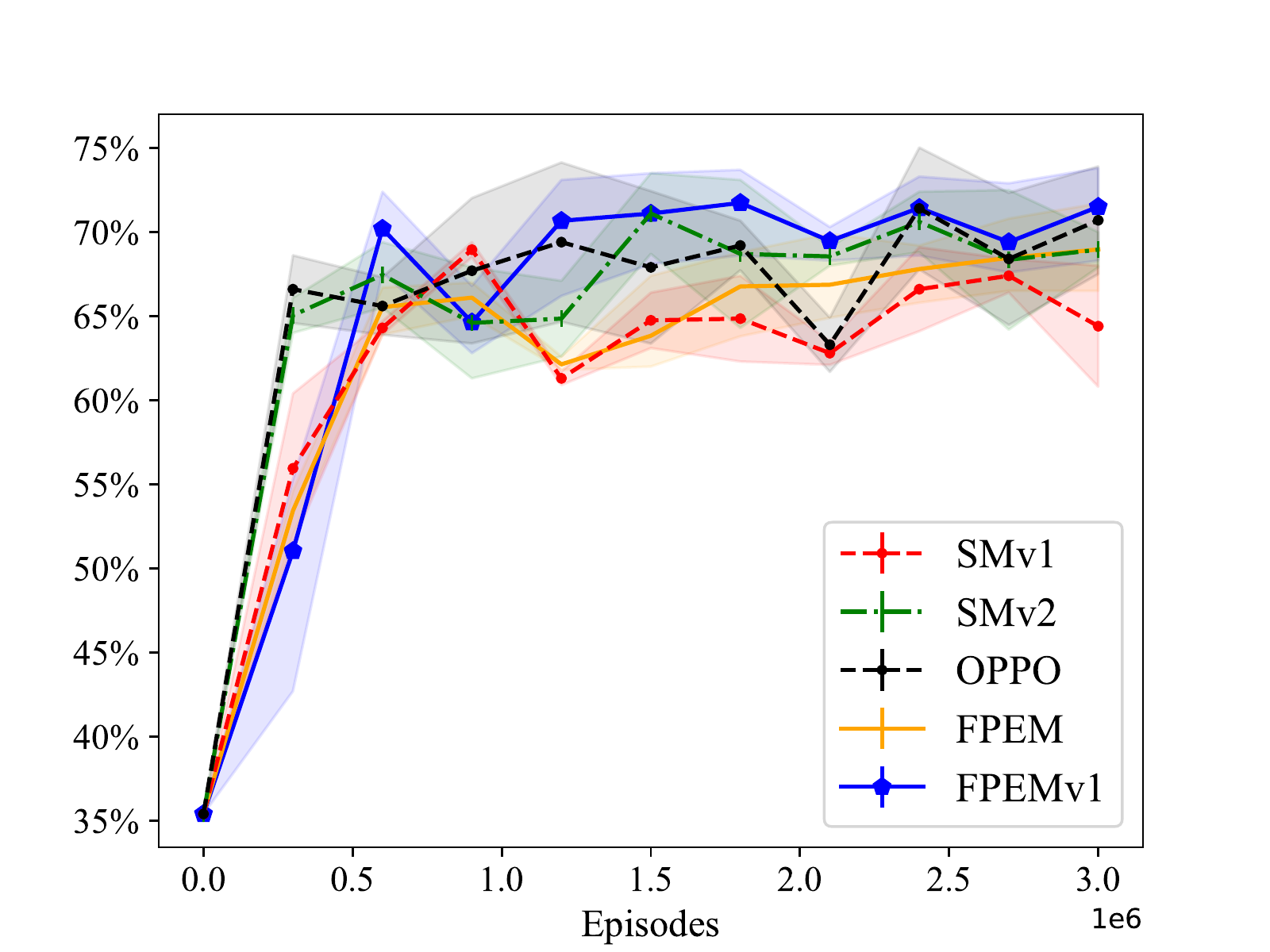}}
	\subfigure[HNR-10]{
		\label{Fig5.sub.4}
		\adjincludegraphics[width=0.48\textwidth, trim={0 {0.0\width} {.0\width} {.0\width}}, clip]{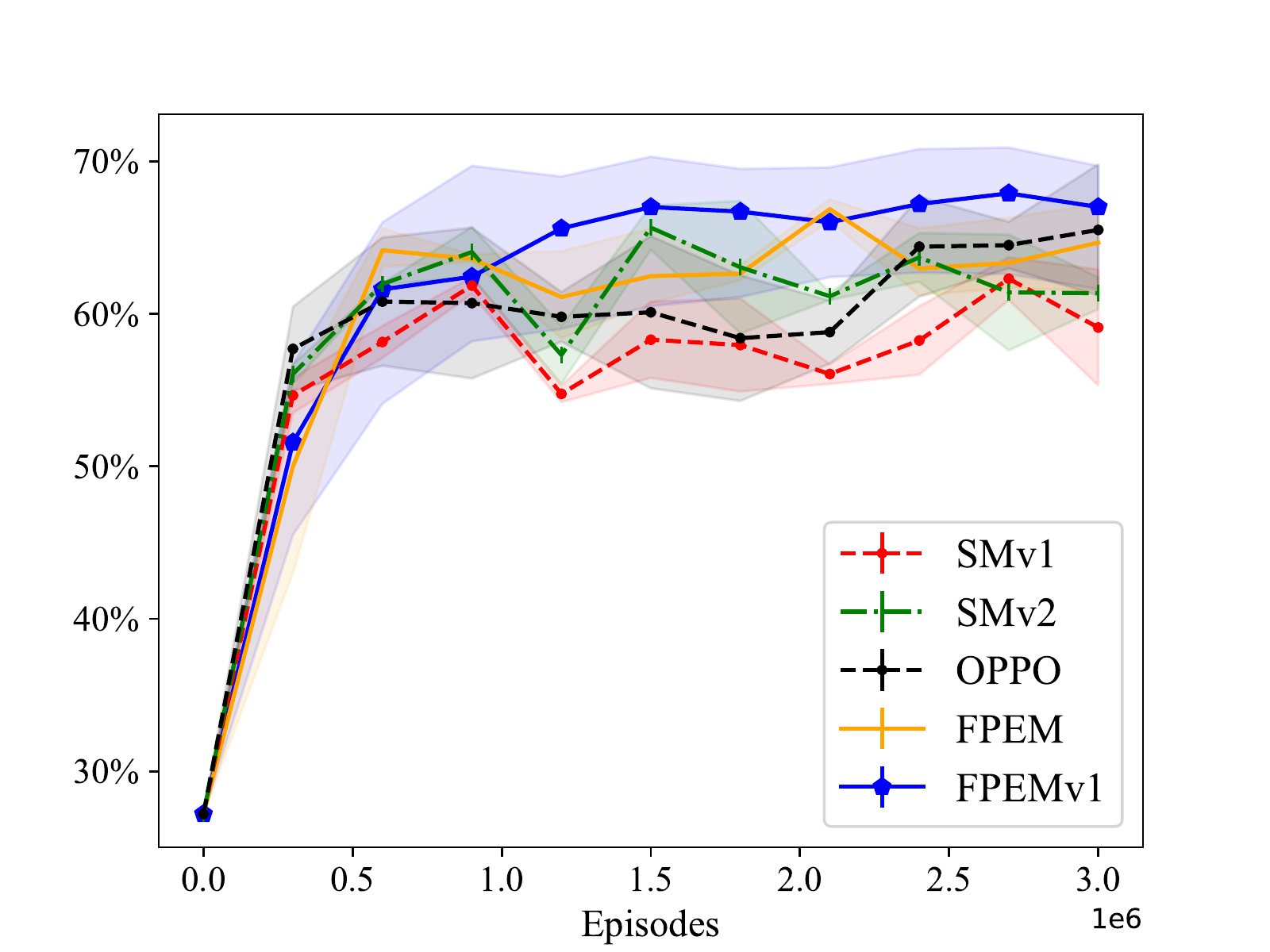}}
	\caption{Average winning rates against different built-in AIs in Mini-RTS. Each point is tested using $2\times 10^4$ episodes. } 
	\label{Fig.rts_ai}
\end{figure*}

In Mini-RTS, only OPPO, SMv1 and SMv2 are compared with FPEM. Having noticed that the two players are homogeneous and use the same policy model in SMv1 and OPPO, these two approaches are trained through self-play, where both players share the model parameters and play against itself. NFSP is omitted due to the slow convergence in treasure hunting, and from \cite{nfsp_rts19}, NFSP falls behind self-play, where the implementation of SMv1 essentially is equivalent to raw self-play in that work. The PSRO is not completely conducted since it spends too much time on computing the expected utility tensor and can not get the Nash distribution. All of the evaluated approaches are trained without built-AIs and make a decision every 50 ticks, and the base policies are initialized by a pre-trained model through self-play for $40K$ episodes. The results against trained adversaries are shown in Table \ref{Tbl.rts}. Each adversary is the corresponding opponent's policy at the end of each min iteration and is evaluated for $2\times 10^4$ episodes and the standard deviation is calculated from the results of 3 runs. Since SMv1 and OPPO are totally self-play, the average winning rates is around 50\%, however, SMv1 achieves a lower standard deviation. So a behavior policy performs more stably. Besides, FPEM, FPEMv1 have a higher overall winning rates against the adversaries.

To further evaluate the performance of FPEM, all the trained models play against two built-in AIs with different configurations: with decision frequencies 50 and 10 ticks (the latter one is tougher). Since the built-in AIs do not learn to adapt, a policy againt a same AI should defeat it once and for all. However, from the results against built-in AIs in Figure \ref{Fig.rts_ai}, all non-expanding models have notable drops when playing against the same AIs as the training proceeds, except FPEM and FPEMv1. That is, without expanding models or a behavior policy, the policy is easy to forget how to respond to old opponents. The shaded areas denote the average winning rates with the standard deviation from 3 runs. Besides, we let FPEMv1 compete with the three approaches, where each match consists of $2\times 10^4$ episodes and only the final models are tested. FPEMv1 defeats OPPO, SMv1, SMv2 with winning rates of 53.5\%, 51.9\%, 52.2\%, with a maximum deviation of 0.5\% respectively.

So, with expanding modes and the behavior-form policy, FPEM is significantly more robust against various opponents, with little or even without a sudden winning rate drop as the number of base policies increases.

\section{Conclusion}
In this paper, we propose fictitious play reinforcement learning with expanding models (FPEM) that consists of multiple sub-models (base policies) and can be used as a behavior policy. The proposed method is scalable with a fixed number of base policies, which alliviates the forgetting problem and can improve the learning efficiency. Since the uniform version of FPEM tracks the process of fictitious play, it enjoys the convergence property in two-player zero-sum games. However, FPEM is not restricted to learn a uniform mixture over the sub-models, and any other distribution can be used. Experimental results on different kinds of domains demonstrate that the learned FPEM model does not forget how to respond to old opponents and is harder to be exploited by malicious opponents, compared to NFSP or a pool of policies.

\section{Acknowledgement}

This work is supported by the NSFC (61876077), Jiangsu SF (BK20170013), and Collaborative Innovation Center of Novel Software Technology and Industrialization.

\bibliographystyle{abbrvnat}
\bibliography{aamas20}

\end{document}